\documentclass[9pt, conference]{IEEEtran}
\IEEEoverridecommandlockouts
\usepackage{cite}
\usepackage{amsmath,amssymb,amsfonts}
\usepackage{graphicx}
\usepackage{textcomp}
\usepackage{xcolor}
\usepackage{caption}
\usepackage{tikz}
\usepackage{array}

\usepackage{hyperref,xcolor}

\usepackage{mdwmath}
\usepackage{mdwtab}
\usepackage{eqparbox}

\usepackage[tight,footnotesize]{subfigure}
\usepackage{amsfonts}
\usepackage{tikz}
\usepackage{textcomp}
\usepackage{wrapfig}
\usepackage{comment}
\usepackage{tcolorbox}

\usepackage{tabularx}
\usepackage{lscape}
\usepackage{mathtools}
\usepackage{booktabs}
\usepackage{algpseudocode}
\usepackage{filecontents}
\usepackage{algorithm}

\usepackage{times}

\usepackage{soul}
\usepackage{url}
\usepackage[utf8]{inputenc}
\usepackage{caption}
\usepackage{amsmath,amssymb,amsfonts}
\usepackage{booktabs}
\usepackage{tikz}
\usepackage{amsmath}

\usepackage{times}

\usepackage{soul}
\usepackage{url}

\usepackage[utf8]{inputenc}
\usepackage{graphicx}
\usepackage{booktabs}
\usepackage{booktabs}
\usepackage{algpseudocode}
\usepackage{algorithm}
\usepackage[none]{hyphenat}
\DeclareMathOperator*{\argmax}{arg\,max}

\def\BibTeX{{\rm B\kern-.05em{\sc i\kern-.025em b}\kern-.08em
    T\kern-.1667em\lower.7ex\hbox{E}\kern-.125emX}}
\begin{document}

\title{ SSAP: A Shape-Sensitive Adversarial Patch for Comprehensive Disruption of Monocular Depth Estimation in Autonomous Navigation Applications 
} 

\author{%
Amira Guesmi$^{1}$, Muhammad Abdullah Hanif$^1$,  
Ihsen Alouani$^2$, Bassem Ouni$^3$, Muhammad Shafique$^1$ \\
$^1$ eBrain Lab, New York University (NYU) Abu Dhabi, UAE \\ 
$^2$ CSIT, Queen’s University Belfast, UK\\
$^3$ AI and Digital Science Research Center, Technology Innovation Institute (TII)\\
}


\maketitle

\begin{abstract}

Monocular depth estimation (MDE) has advanced significantly, primarily through the integration of convolutional neural networks (CNNs) and more recently, Transformers. However, concerns about their susceptibility to adversarial attacks have emerged, especially in safety-critical domains like autonomous driving and robotic navigation. Existing approaches for assessing CNN-based depth prediction methods have fallen short in inducing comprehensive disruptions to the vision system, often limited to specific local areas.
In this paper, we introduce SSAP (Shape-Sensitive Adversarial Patch), a novel approach designed to comprehensively disrupt monocular depth estimation (MDE) in autonomous navigation applications. Our patch is crafted to selectively undermine MDE in two distinct ways: by distorting estimated distances or by creating the illusion of an object disappearing from the system's perspective. Notably, our patch is shape-sensitive, meaning it considers the specific shape and scale of the target object, thereby extending its influence beyond immediate proximity.
Furthermore, our patch is trained to effectively address different scales and distances from the camera. Experimental results demonstrate that our approach induces a mean depth estimation error surpassing 0.5, impacting up to 99\% of the targeted region for CNN-based MDE models. Additionally, we investigate the vulnerability of Transformer-based MDE models to patch-based attacks, revealing that SSAP yields a significant error of 0.59 and exerts substantial influence over 99\% of the target region on these models.

\end{abstract}
\maketitle
\section{Introduction}\label{sec:intro}
\label{sec:intro}

\begin{figure*}[!ht]
\centering
\includegraphics[width=1.7\columnwidth]{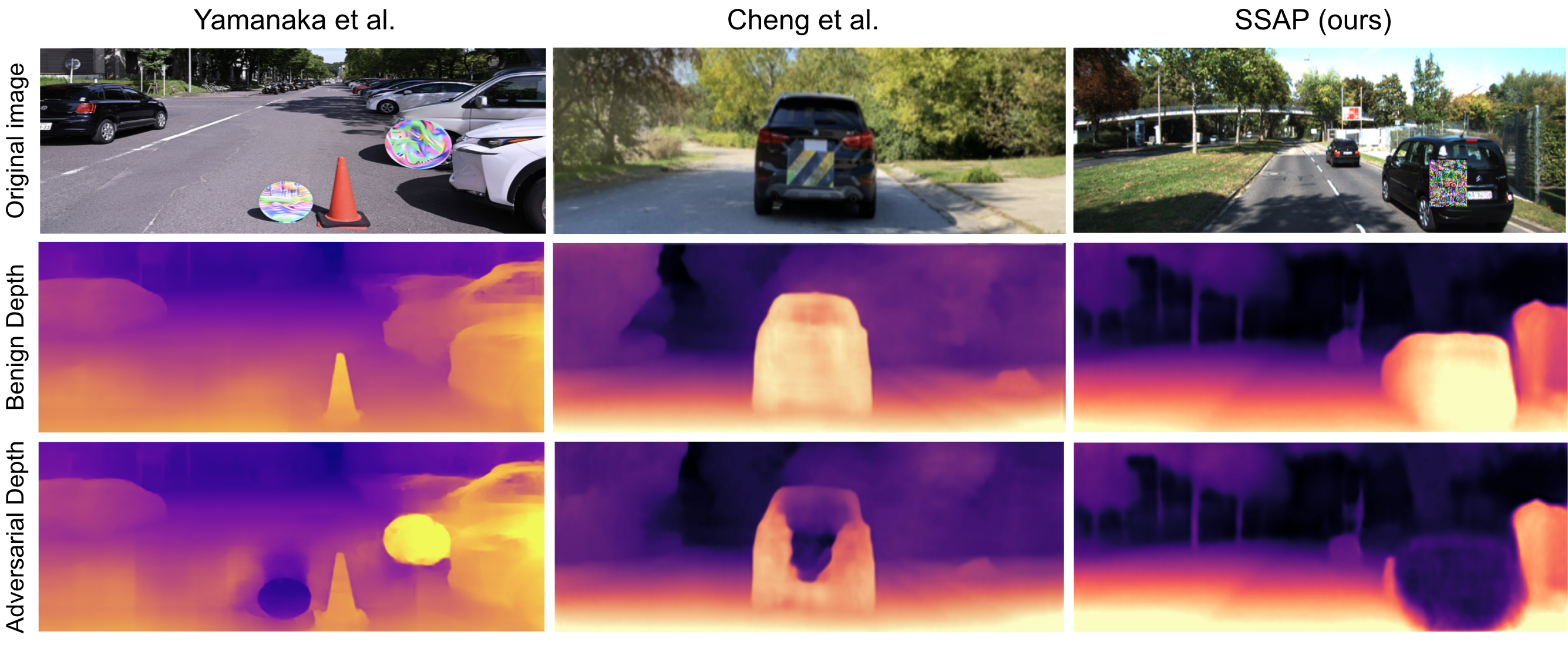}
\caption{Our SSAP makes the object fully disappear, in contrast, adversarial patches proposed by Yamanaka et al. \cite{Yamanaka_Access} and Cheng et al. \cite{Cheng_ECCV} are weak adversarial patches that only impact the depth of a small region of the target object which is restricted to the overlapping region between the patch and the input image.}
\label{results}
\end{figure*}

MDE has become increasingly valuable in practical applications such as robotics and autonomous driving (AD). MDE enables the extraction of depth information from a single image, improving understanding of the scene. Its importance extends to critical robotic functions, including obstacle avoidance \cite{obstacles}, object detection \cite{autonomous2}, visual SLAM \cite{robotics1, slam}, and visual relocalization \cite{relocalization}.

Various methods for depth estimation rely on technologies such as RGB-D cameras, Radar, LiDAR, or ultrasound devices to directly capture depth information within a scene. However, these alternatives have significant limitations. RGB-D cameras have a limited measurement range, while LiDAR and Radar provide sparse data and are costly sensing solutions. These factors may not be suitable for compact autonomous systems, such as low-cost, lightweight, and small-sized mobile robots. Additionally, ultrasound devices suffer from inherent measurement inaccuracies. Furthermore, these technologies consume substantial energy and have large form factors, making them unsuitable for resource-restricted, small-scale systems that must adhere to stringent real-world design constraints. In contrast, RGB cameras offer lightweight and cost-effective options. They have the capability to provide more comprehensive environmental data.

Leading players in the autonomous vehicle sector are driving advancements in self-driving technology by utilizing cost-effective camera solutions. Notably, Monocular Depth Estimation (MDE) has been seamlessly integrated into Tesla's production-grade Autopilot system \cite{tesla2, tesla}. Other major autonomous driving (AD) enterprises, such as Toyota \cite{toyota} and Huawei \cite{huawei}, are also adopting this approach to accelerate self-driving advancements, following Tesla's lead.

In recent years, the advancement of deep learning has significantly improved the performance of monocular depth estimation (MDE), primarily through the utilization of CNN-based models \cite{fu2018deep, song2021monocular, huynh2020guiding} and Transformer-based models \cite{chang2021transformer, varma2022transformers, mimdepth}. However, CNNs have shown vulnerabilities to adversarial attacks, and the security properties of Transformers are yet to be thoroughly studied.

Previous efforts in patch-based adversarial attacks, which focused solely on CNN-based MDE \cite{Yamanaka_Access, Cheng_ECCV, saam}, produced relatively weak adversarial patches. These patches had a limited impact on the depth estimation of specific objects such as vehicles and pedestrians, with their influence confined to the overlapping area between the patch and the input image. There is significant potential for improvement in expanding the sphere of influence of these patches.

Our approach introduces a novel technique for crafting adversarial patches tailored for both CNN-based and Transformer-based monocular depth estimation. Specifically, we generate shape-sensitive adversarial patches aimed at deceiving the target methods. These patches prompt the methods to erroneously estimate the depth of designated objects (such as vehicles or pedestrians), or even to completely conceal the presence of those objects.


\textbf{In summary, the \textit{novel contributions} of this work are:}
\begin{itemize}
\item We introduce a shape-sensitive adversarial patch (SSAP) designed to disrupt the output of the MDE model.
\item We leverage information from a pre-trained detector during the patch generation process. This enables us to adaptively craft patches that are robust to varying scales and distances from the camera, mimicking real-world scenarios. Additionally, this approach helps prevent the patch from being trained for irrelevant objects, ensuring that it exclusively targets specific objects and thereby enhances its effectiveness.
\item We introduce a novel penalized loss function aimed at enhancing the efficiency of our adversarial patch and expanding its impact region (refer to Figure \ref{results}).
\item We conduct an ablation study to demonstrate the effectiveness of our modified loss function in extending the influence of our proposed patch.
\item To the best of our knowledge, we are the first to investigate the robustness of transformer-based MDE models. We demonstrate their vulnerability to our patch-based adversarial attacks, despite claims of robustness to natural noise and adversarial attacks \footnote{We provide a demo video showcasing the effectiveness of our patch in concealing a target object for the transformer-based MDE model MIMDepth \cite{mimdepth}. The video demonstrates various patch sizes and distances from the camera.}.
\item Our proposed patch achieves a high mean depth estimation error exceeding 0.5, significantly impacting nearly 99\% of the target region for CNN-based MDE. Additionally, it results in a mean depth estimation error of 0.59 with a substantial influence on 99\% of the region for transformer-based MDE.
\item Our devised attack methodology is generic, making it applicable to various object categories present on public roads. However, for proof-of-concept, we focus on two representative object types —cars and pedestrians— for targeting purposes.
\end{itemize}
An overview of our framework is depicted in Figure \ref{approach}, while a comprehensive description can be found in Section \ref{overview}.   

\section{Proposed Approach}\label{sec:method}
\subsection{Problem formulation}
In monocular depth estimation, when presented with a benign image denoted as $I$, the objective of the adversarial attack is to cause the depth estimation method to inaccurately predict the depth of the intended object by employing a strategically designed image represented as $I^*$. Technically, the adversarial example incorporating the generated patch can be mathematically formulated as follows:

\begin{equation}
   I^* = (1 - M_P) \odot I + M_P \odot P
\end{equation}

$\odot$ is the component-wise multiplication, denoted as $P$, with the specific property, and $M_P$ is the patch mask, used to constrict the size, shape, and location of the adversarial patch. 
The adversarial depth, i.e., the output of the victim model $F$ when taking as input the adversarial example is:
\begin{equation}
  d_{adv} = F((1 - M_P) \odot I + M_P \odot P)
\end{equation}
The problem of generating an adversarial example can be formulated as a constrained optimization \ref{eq:adv}, given an original input image $I$ and a MDE model $ F(.) $,:
\begin{equation}
\label{eq:adv}
   arg\min_{P} \left\|P\right\|_p  \\
   s.t. F((1 - M_P) \odot I + M_P \odot P) \neq F(I)\\
    d_{adv} \neq d
\end{equation}

The goal is to identify a minimal adversarial noise, $P$, such that when applied on any object within a designated input domain $U$, it strategically compromises the underlying DNN-based MDE model $F(.)$. This compromise can take the form of either distorting the estimated distance or causing the object to vanish from the prediction.

It's important to note that an analytical solution isn't feasible for this optimization task due to the non-convex nature of the DNN-based model $F(.)$ involved. As a result, the formulation of Equation \ref{eq:adv} can be expressed as follows, allowing for the utilization of empirical approximation techniques to numerically address the problem:
\begin{equation}
\label{eq:formulation}
    \argmax_{P} \sum_{I \in \mathcal{U}} l(F((1 - M_P) \odot I + M_P \odot P), F(I))
\end{equation}
Here, $l$ represents a predefined loss function, and $\mathcal{U} \subset U$ denotes the attacker's training dataset. To tackle this challenge, optimization methods such as Adam \cite{adam} can be employed to address the problem. During each iteration of the training process, the optimizer iteratively updates the adversarial patch $P$.

\subsection{Overview}
\label{overview}
While our approach does not directly incorporate considerations of stealthiness, we have focused on ensuring a meaningful and fair comparison with existing methods. To achieve this, we opted to replicate the methodology outlined by Cheng et al. \cite{Cheng_ECCV}, While not specifically addressing stealthiness. Furthermore, unlike Cheng et al.'s method, which involves training a patch for consistent placement on the same object at a fixed distance from the camera while only changing the background, a scenario that may not accurately reflect real-world conditions, our method ensures that the generated patch remains effective across various scales and distances from the camera. This adaptability to diverse conditions enhances its practical applicability and effectiveness in real-world settings. Additionally, we replicate the methodology presented in \cite{Yamanaka_Access}, wherein a patch was trained for arbitrary placement within the scene.

As illustrated in Figure \ref{results}, our SSAP achieves the complete disappearance of the targeted object, while the other two patches exhibit a less significant impact. To ensure consistency in our comparison, we maintained uniformity by utilizing identical optimization parameters, patch dimensions, and shapes.

Our study aims to develop adversarial patches with a broad impact, covering the entire object regardless of its size, shape, or position, while maintaining their effectiveness as attack tools. To achieve this goal, we utilize a pre-trained object detector to accurately pinpoint the location of the targeted object—the object we aim to hide or manipulate—along with its predicted depth. This precise identification enables us to create a patch that remains effective even with varying distances between the camera and the object.

\begin{figure}[!ht]
\centering
\includegraphics[width=1\columnwidth]{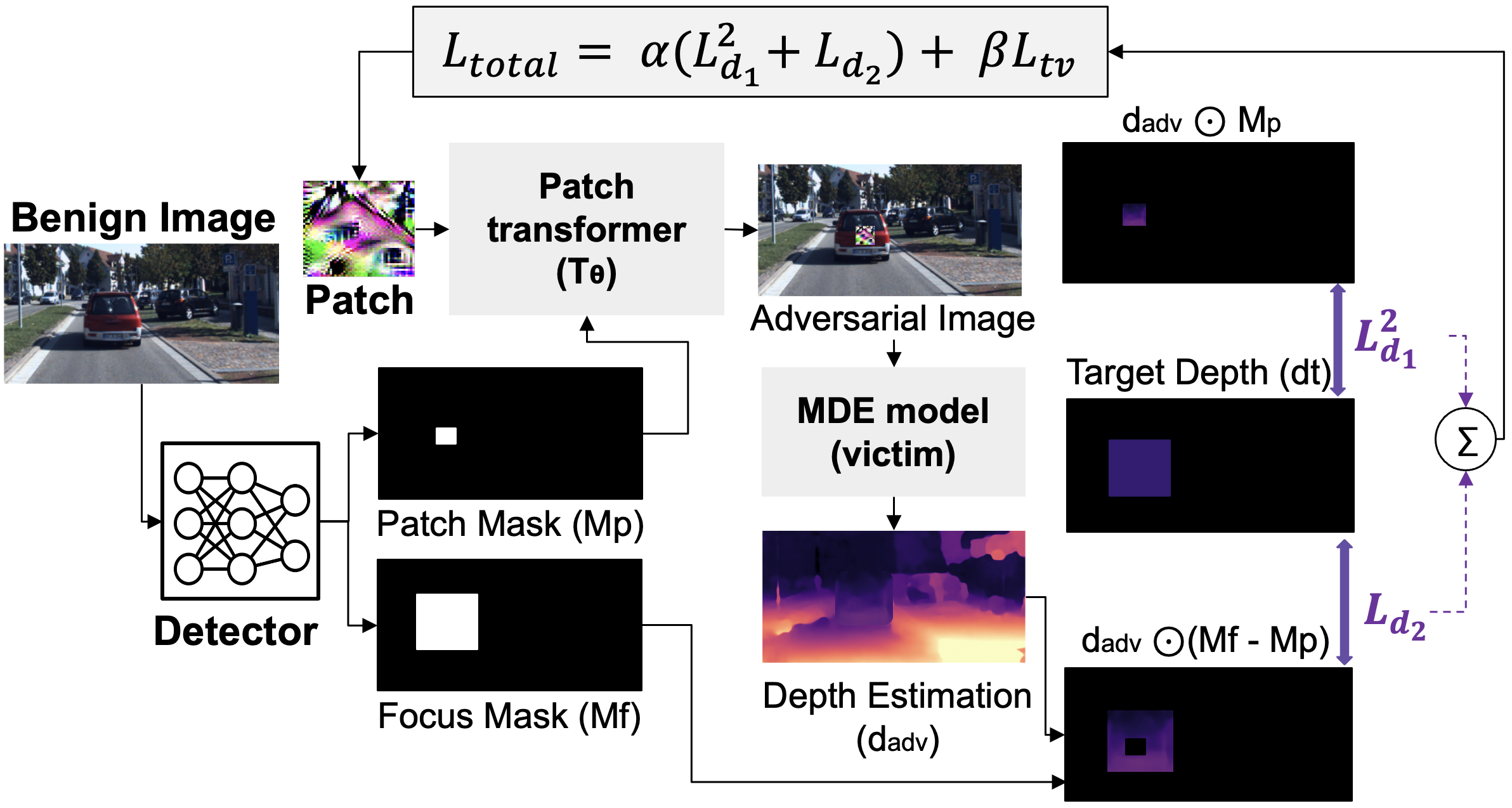}
\caption{Overview of the proposed approach. 
}
\label{approach}
\end{figure}

Furthermore, we introduce a novel loss function designed to amplify the patch's effect and expand the area it influences. As shown in Figure \ref{approach}, we begin with a pre-trained object detector and initiate the process by creating two distinct masks. The first mask, denoted as $M_p$, represents the precise placement of the patch at the center of the target object. The second mask, $M_f$, corresponds to the specific region occupied by the target object—essentially, the area influenced by the target object. Subsequently, the patch is fed to the patch transformation block, also known as the 'patch transformer', where we implement the geometric alterations outlined in Section \ref{GT}.

After applying these transformations, we utilize 'the patch applier' to superimpose the generated patch onto the input image. This process involves incorporating information obtained from the object detector, as explained in Section \ref{PA}. Following this step, we perform a forward pass. In the subsequent stage, we use the generated masks to compute the required loss functions. Then, we calculate the gradient of the patch. This gradient information is used to update the actual patch, denoted as $P$.


\subsection{Patch Applier}
\label{PA}
In a physical setting, the attacker's control over the perspective, scale, and positioning of the patch in relation to the camera is limited. Therefore, we aim to enhance the robustness of our patch to accommodate a wide range of potential scenarios. During the patch generation phase, we overlay the patch onto the surface of the target object, such as the rear of a vehicle or human attire. This methodology enables us to simulate diverse scenes with different settings.

The training process involves a variety of transformations, such as rotations and occlusions, carefully incorporated to simulate the plausible appearance of our adversarial patch $P$ in a realistic context. Subsequently, leveraging information from the object detector, we obtain precise object locations (such as vehicles or individuals) within a given image $I$. At this stage, it becomes feasible to place our adversarial patch $P$ onto the identified object. Two distinct masks arise from this process: 
$M_f$, encircling the object to demarcate its influence, and $M_p$, designed to restrict the patch's characteristics, including its location, dimensions, and shape. It is crucial to emphasize that the object detector's role is limited to the optimization procedure of the patch and is not extended to the actual attack phase. To eliminate the need for manual patch placement onto objects as done in \cite{Cheng_ECCV}, we utilize the YOLOv4-tiny detector pretrained on the MSCOCO dataset \cite{Lin2014MicrosoftCC}.

The detector's capabilities streamline the process by automatically identifying object placements, contributing to a more efficient and effective workflow. Let $U = \{I_i\}_{i=1}^M$ and $V = \{J_j\}_{j=1}^N$ respectively be the $M$ training and $N$ testing images for a particular scene of attack. We run the YOLOv4-tiny object detector on $U$ and $V$ with an objectness threshold of 0.5 and non-max suppression IoU threshold of $0.4$. This yields the tuples:
\begin{equation}
    T_i^U = \{(B_{i,k}^U)\}_{k=1}^{D_i} , T_j^V = \{(B_{j,l}^V)\}_{l=1}^{E_j}
\end{equation}

for each $I_i$ and $J_j$ , where $D_i$ is the number of detections in $I_i$ (fixed to 14 in our experiment same as in \cite{naturalistic}). and $B_{i,k}^U$ is the bounding box of the $k-th$ detection in $I_i$. 
(Same for $E_j$ and $B_{j,l}^V$ for $J_j$). 
The sets of all detected objects are:
\begin{equation}
    T^U = \{T_i^U \}_{i=1}^M , T^V = \{T_j^V \}_{j=1}^N
\end{equation}

The information used to optimize $P$ for a scene of attack are the training images $U$ and annotations $T^U$. The patch $P$ is randomly initialized. Given the current P, the patch is rendered on top of each detected object of the chosen class for each training image $I_i$ using the mask $M_{p_{i,k}}^U$. 
$M_{p_{i,k}}^U$ is a matrix of zeros except for the patch location (The center of the patch is the center of the bounding boxes).

The focus mask $M_{f_{i,k}}^U$ is defined as the space limited by the generated bounding boxes $B_{i,k}^U$ in a way that it covers the whole detected object. $M_{f_{i,k}}^U$ is a matrix of zeros except for the targeted regions (objects covered area) where the pixel values are ones. We multiply the generated masks with the adversarial depth map $d_{adv}$ and use the resulting maps to compute the two losses.

\subsection{Patch Transformation Block}
\label{GT}

The positioning and perspective of a camera on an autonomous vehicle in relation to another vehicle or target object undergo continuous variation. The images captured and provided to the victim model are taken from various distances, angles, and lighting conditions. Therefore, any modification introduced by an attacker, such as an adversarial patch, must be resilient to these evolving circumstances. To simulate this variability, a range of physical transformations is applied, each representing different conditions that may occur. These transformations include introducing noise, applying random rotations, altering scales, and simulating variations in lighting. The patch transformer is utilized to implement these transformations effectively.


The transformations executed include:

Random Scaling: The patch's dimensions are randomly adjusted to approximately match its real-world proportions within the scene.
Random Rotations: The patch $P$ is subjected to random rotations (up to $\pm20^\circ$) centered around the bounding boxes $B_{i,k}^U$. This emulates uncertainties related to patch placement and sizing during printing.
Color Space Transformations: Pixel intensity values are manipulated through various color space transformations. These include introducing random noise (within $\pm0.1$ range), applying random contrast adjustments (within the range $[0.8, 1.2]$), and introducing random brightness adjustments (within $\pm0.1$ range).

\subsection{Penalized Depth Loss}
\label{newloss}

The bounding boxes $B_{i,k}^U$ generated serve as the foundation for creating a focus mask $M_f$, which encompasses the specific region where we intend to modify the predicted depth.
Our objective is to extend the region influenced by the patch, going beyond mere pixel overlap. To achieve this, we decompose the depth loss $L_{depth}$ into two distinct terms: $L_{d_1}$ and $L_{d_2}$. $L_{d_1}$ represents the loss incurred by pixels that are overlapped by the patch, while $L_{d_2}$ pertains to the loss stemming from pixels that don't overlap.

To direct the optimization process towards prioritizing the reduction of the non-overlapped pixel loss $L_{d_2}$, we employ a squaring operation on the term denoting the disparity between the output depth and the target depth, denoted as $\left | d_t - d_{adv} \right | \odot M_P$. This utilization of quadratic functions is strategic, as these functions exhibit a slower rate of increase (slope or rate of change), consequently delaying the convergence of overlapped pixel loss in comparison to non-overlapped pixels.
The losses are defined as the distance between the predicted depth and the target depth and calculated as follows:
\begin{equation}
    L_{d_1} = \frac{1}{m \times n} \sum_{i,j} \left | d_t - d_{adv} \right | \odot M_P 
\end{equation}

\begin{equation}
    L_{d_2} = \frac{1}{m \times n} \sum_{i,j} \left | d_t - d_{adv} \right |\odot (M_f - M_P)
\end{equation}

\begin{equation}
    L_{depth} =  L_{d_1}^2 + L_{d_2} 
\end{equation}


\subsection{Adversarial Patch Generation}
We iteratively perform gradient updates on the adversarial patch $(P)$ in the pixel space in a way that optimizes our objective function defined as follows:

\begin{equation}
    L_{total} = \alpha L_{depth} + \gamma L_{tv} 
\end{equation}

$L_{depth}$ is the adversarial depth loss.
$L_{tv}$ is the total variation loss on the generated image to encourage smoothness \cite{mahendran2015understanding}.
It is defined as:
\begin{equation}
    L_{tv} = \sum_{i,j} \sqrt{(P_{i+1,j} - P_{i,j})^2 + (P_{i,j+1} - P_{i,j})^2}
\end{equation}

where the sub-indices $i$ and $j$ refer to the pixel coordinate of the patch $P$. $\alpha$ and $\beta$ are hyper-parameters used to scale the three losses. For our experiments, we set $\alpha =1 $ and $\beta = 2$. 
We optimize the total loss using Adam \cite{adam} optimizer. We try to minimize the object function $L_{total}$ and optimize the adversarial patch. We freeze all weights and biases in the depth estimator and only update the pixel values of the adversarial patch. The patch is randomly initialized. 

\section{Experimental Results}\label{sec:exp}
In our experimental setup, we employ four MDE models: Three CNN-based models, namely, monodepth2 \cite{monodepth2}, Depthhints \cite{Depthhints}, and Manydepth \cite{Manydepth}, along with the Transformer-based MIMdepth \cite{mimdepth}.
These models were chosen based on their practicality and the availability of open-source code. It's worth noting that the first three models are the same ones featured in the work presented in \cite{Cheng_ECCV}.

For our evaluations, we use real-world driving scenes extracted from the KITTI 2015 dataset \cite{kitti}. This dataset comprises synchronized stereo video recordings alongside LiDAR measurements, all captured from a moving vehicle navigating urban surroundings. The dataset encapsulates an extensive array of road types, including local and rural roads as well as highways. Within these scenes, a diverse range of objects is present, such as vehicles and pedestrians, allowing for a comprehensive evaluation of the attack performance. Additionally, we employ the CASIA datasets \cite{casia} to test person objects.

We train our patch on a TeslaV100 GPU. Our patch optimization process is executed for $500$ epochs, using the Adam optimizer with a learning rate set at $0.01$. The patch is scaled to a factor of $0.2$ during this process.

\subsection{Evaluation Metrics}
To assess the efficacy of our proposed attack, we utilize same metrics as employed in \cite{Cheng_ECCV}: the \textit{mean depth estimation error} ($E_d$) attributed to the target object, and the \textit{ratio of the affected region} ($R_a$). To compute these metrics, we define the depth prediction of the clean target object as the ground truth. 

The \textit{mean depth estimation error} quantifies the degree to which our proposed patch impacts the accuracy of depth estimation. A higher value in this metric indicates a more effective attack. Similarly, in the \textit{ratio of the affected region}, a higher value indicates improved attack performance. 
In contrast to the approach presented in \cite{Cheng_ECCV}, where a patch is trained on a single object situated at a fixed distance from the camera while varying only the background, our approach involves training the patch for diverse distances. This includes objects positioned at various locations relative to the camera, such as close, far, left, right, or center.
Initially, the output of the disparity map is normalized within the range of $0$ to $1$, where $0$ signifies the farthest point in the image and $1$ represents the closest point. 

\textit{The mean depth estimation error} was measured using the following metrics:
\begin{equation}
    E_{d} = \frac{\sum_{i,j}(\arrowvert d - d_{adv} \arrowvert \odot M_f) }{\sum_{i,j}{M_f }}
\end{equation}


The \textit{ratio of the affected region} is determined by evaluating the percentage of pixels whose depth values have been modified beyond a certain threshold. This evaluation is performed concerning the total number of pixels covered by the focus mask ($M_f$). 
For objects with any alteration in a pixel's depth value surpassing $0.1$ results in that pixel being considered as affected. 

The \textit{ratio of the affected region} was measured using the following metrics:
\begin{equation}
    R_{a} = \frac{\sum_{i,j}\textbf{I}((\arrowvert d - d_{adv} \arrowvert \odot M_f) > 0.1) }{\sum_{i,j}{M_f }}
\end{equation}


Additionally, we employ the Mean Square Error (MSE) to assess the performance of the model in relation to the predicted output depth map derived from an unperturbed input. The formulation of this metric is provided below.
\begin{equation}
    MSE = \frac{1}{N}\sum_{i,j}(d_{adv_{i,j}} - d_{i,j})^2  
\end{equation}

where $N$ is the total number of pixels.
\subsection{Evaluation Results}

We extend our attack evaluation to include the four MDE models, targeting both classes of objects for each model. Initially, we generate adversarial patches specifically tailored for pedestrians and cyclists. In this phase, the patch scale is maintained at $0.2$ for consistent testing across the different models.
\begin{figure}[!ht]
\centering
\includegraphics[width=\columnwidth]{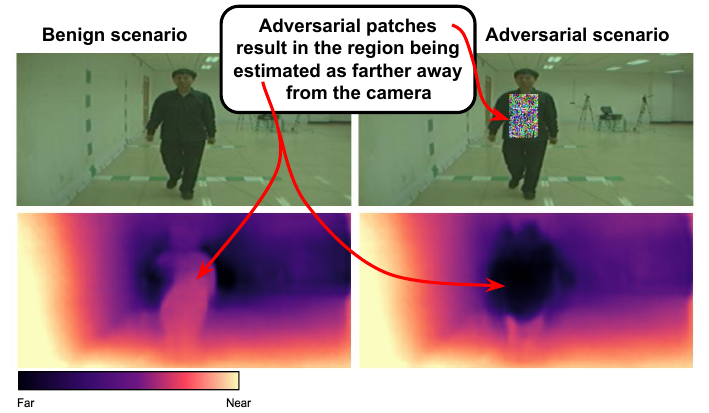}
\caption{Impact of SSAP on person/pedestrian class for Transformer-based MDE. The person seamlessly blends with the background.}
\label{results_ped}
\end{figure}
As depicted in Figure \ref{results_ped}, our patch exhibits remarkable effectiveness by completely concealing the person. Intuitively, smaller objects are relatively easier to hide or manipulate in terms of their depth. However, achieving this for smaller objects requires smaller patches, which consequently result in a relatively reduced impact.

In the following experiment, we proceed to create an adversarial patch targeting the "car" class. Utilizing a patch scale of $0.2$, we observe that nearly all objects integrated with the SSAP achieve complete concealment. This outcome remains consistent regardless of the object's specific characteristics, including factors such as shape, size, and proximity to the camera. The success in concealing various objects corroborates the robustness of our proposed patch.
\begin{figure}[!ht]
\centering
\includegraphics[width=1\columnwidth]{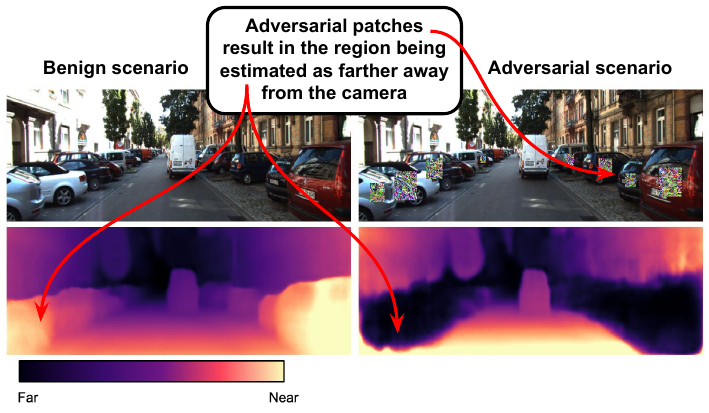}
\caption{Impact of SSAP on car class for Transformer-based MDE.}
\label{results_car}
\end{figure}

We quantitatively assess the performance of our patch using the mean square error $MSE$, the mean depth estimation errors, denoted as $E_{d}$, alongside the ratios of the affected regions, designated as $R_{a}$, for the target object across 100 scenes extracted from the KITTI dataset. The presented results are determined by calculating the average values of these metrics, offering a representative result.
\begin{table}[!htp]
  \centering
\small
  \caption{Attack performance in terms of mean depth estimation error ($E_{d}$) and ratio of affected region ($R_{a}$).} 
  \label{performance}
  \begin{tabular}{lccc}
    \toprule
    \textbf{Models} & \textbf{$MSE$} & \textbf{$E_{d}$} & \textbf{$R_{a}$} \\
    \midrule
      \textbf{Monodepth2}  & 0.49  & 0.55 & 0.99 \\
      \textbf{Depthhints}  & 0.47 & 0.53 &  0.98 \\
      \textbf{Manydepth}   & 0.46 & 0.53  & 0.98 \\
      \textbf{MIMdepth}   &  0.53  &  0.59  & 0.99 \\
  \bottomrule
\end{tabular}
\end{table}

As reported in Table \ref{performance}, when our patch targets a Transformer-based model (MIMdepth), it achieves an average alteration of approximately 59\%. Considering a scenario set on a highway, where the speed limit stands at 160 km/h, the recommended safe distance extends to 96 meters. When we incorporate our patch, a car positioned 90 meters away is projected to be situated approximately 143.1 meters away. This starkly demonstrates the profound consequences of our devised attack.
Additionally, it's noteworthy that our patch yields an impact on over 99\% of the target region. This data underscores the extensive impact our patch has on altering depth perception.

\begin{figure*}[!ht]
\centering
\includegraphics[width=2\columnwidth]{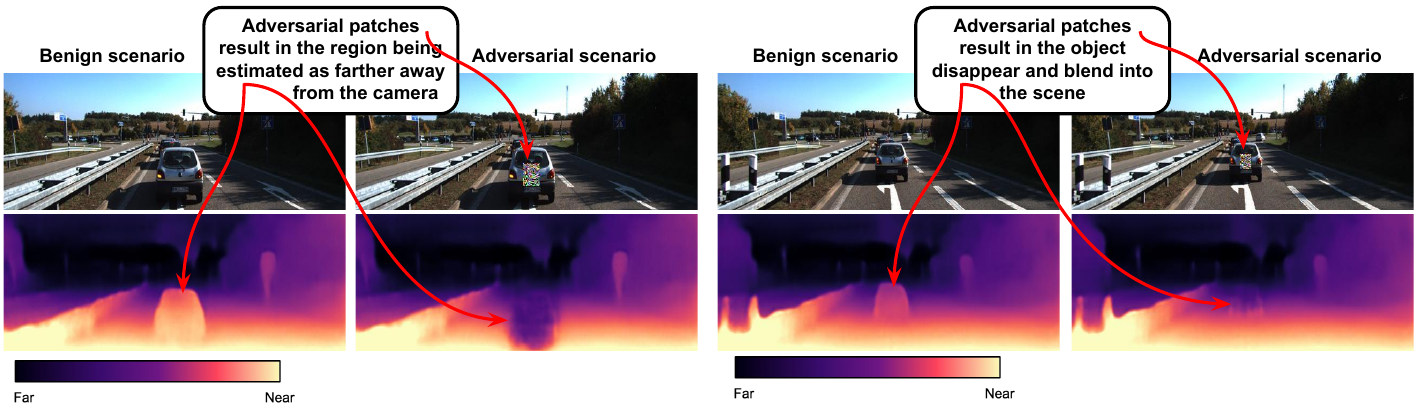}
\caption{Impact of SSAP on car class for transformer-based MDE for different distances from the camera.}
\label{results_car_mimdepth2}
\end{figure*}
Moreover, even when targeting MIMdepth, a method touted for its enhanced robustness through the incorporation of transformer architecture and masked image modeling (MIM), our adversarial patch continues to exhibit effectiveness, causing a notable depth estimation error of $0.59$. Figure \ref{results_car_mimdepth2} visually demonstrates the influence of our adversarial patch on the depth prediction of the MIMdepth model. For more qualitative results related to this model, please refer to the supplementary material.
\subsection{SSAP vs. Existing Attacks}
We conducted a series of experiments to provide a quantitative comparison between our attack strategy and prior approaches \cite{Cheng_ECCV} and \cite{Yamanaka_Access}. To carry out this comparison, we employed the monodepth2 model to evaluate both the depth estimation error and the ratio of the affected region. The results presented in Table \ref{performance} demonstrate that our proposed attack consistently achieves the most substantial alteration in depth estimation and encompasses the largest affected region when compared to the referenced prior works.
\begin{table}[!htp]
  \centering
\small
  \caption{SSAP performance vs. existing attacks.}
  \label{performance}
  \begin{tabular}{lccc}
    \toprule
    \textbf{Attack} & \textbf{$E_{d}$}  & \textbf{$R_{a}$}  & \textbf{$MSE$}\\
    \midrule
      \textbf{SSAP}         &  \textbf{0.55} & \textbf{0.99}  & \textbf{0.49}\\
      \textbf{Cheng et al. \cite{Cheng_ECCV}}     &  0.21 & 0.47 &  0.12\\
      \textbf{Yamanaka et al \cite{Yamanaka_Access}}   &  0.13 & 0.26  & 0.05\\
  \bottomrule
\end{tabular}
\end{table}




\section{Discussion}\label{sec:discussion}
\subsection{Ablation Study}

To evaluate the effectiveness of the proposed penalized depth loss, we conducted an ablation study utilizing the monodepth2 model as our target monocular depth estimation (MDE) model. We tested various combinations of loss terms and present the results in Table \ref{ablation}. Our proposed loss yielded the highest depth estimation error, with a value of $0.55$, compared to $0.24$ for the conventional loss and $0.18$ when utilizing \(L_{d_2}\) for the depth loss. Furthermore, our approach resulted in the highest ratio of affected regions, with a value of $0.99$, compared to $0.53$ and $0.36$ for the other combinations.

\begin{figure}[!ht]
\centering
\includegraphics[width=1\columnwidth]{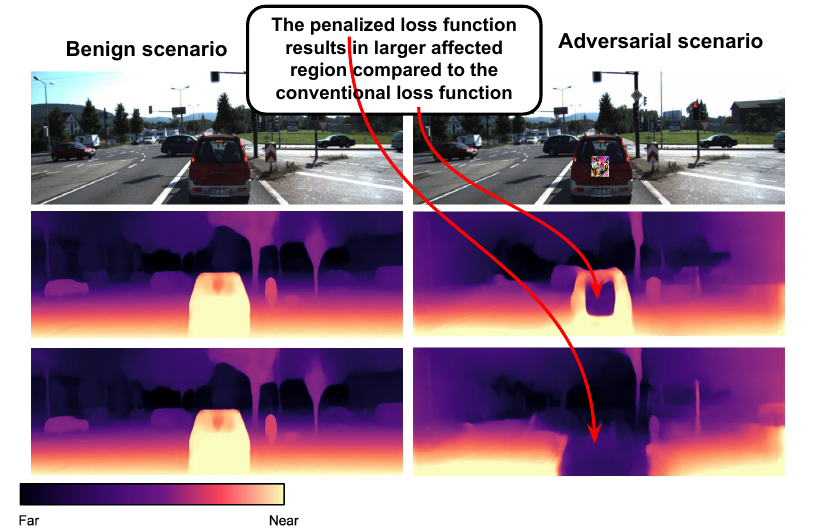}
\caption{Depth prediction w/o the penalized loss function: (Top) the input images, (Middle) results without the penalized depth loss (i.e., $L_{d_1}$ and $L_{tv}$), (Bottom) results with our proposed depth loss (i.e., $L_{d_1}$, $L_{d_2}$ and $L_{tv}$).}
\label{results_without}
\end{figure}

As illustrated in Figure \ref{results_without}, the area affected by the generated patch when using only the $L_{d_1}$ depth is limited to the immediate vicinity of the patch itself. This outcome demonstrates a slightly improved performance compared to the patches featured in \cite{Yamanaka_Access, Cheng_ECCV}.
Subsequently, we proceed to evaluate the effects of our proposed depth loss outlined in Section \ref{newloss}. Upon applying this penalized depth loss, the resulting patch effectively conceals the entire object, as confirmed by our experimentation.

\begin{table}[!htp]
  \centering
\small
  \caption{Attack performance in terms of mean depth estimation error ($E_{d}$) and ratio of affected region ($R_{a}$) for different losses combinations.} 
  \label{ablation}
  \begin{tabular}{lccccc}
    \toprule
    \textbf{$L_{d_1}$} & \textbf{$L_{d_2}$} & \textbf{$L_{tv}$} & \textbf{$E_{d}$} & \textbf{$R_{a}$} \\
    \midrule
                 & \checkmark & \checkmark & 0.18 & 0.36 \\
      \checkmark  &            & \checkmark & 0.24 & 0.53 \\
       \checkmark & \checkmark & \checkmark & 0.55 & 0.99 \\
  \bottomrule
\end{tabular}
\end{table}

\vspace*{-5mm}
\subsection{The Influence of Patch Scale}
We evaluate our attack by targeting the "car" object class using three distinct patch sizes: $0.1$, $0.2$, and $0.3$. We assess the mean depth error and the ratio of the affected region using the three depth estimation models. Table \ref{edc} presents the results for the mean depth estimation error. It is noteworthy that there is a noticeable trend where $E_d$ consistently increases with the increment in patch size across all target models. This outcome aligns with expectations, as larger patches exert a more pronounced influence on the resulting depth error. Similarly, this trend is echoed in the ratio of affected regions, as depicted in Table \ref{rac}, where larger patches correspond to larger affected regions.

\begin{table}[!htp]
  \centering
\small
  \caption{$E_{d}$ for different patch scales.}
  \label{edc}
  \begin{tabular}{cccc}
    \toprule
    \textbf{Scale} & \textbf{Monodepth2}& \textbf{Depthhints}& \textbf{Manydepth}\\
    \midrule
      0.1           & 0.46   & 0.37  & 0.3   \\
      0.2           &  0.55  & 0.53  & 0.53   \\
      0.3           &  0.66 &  0.64 &  0.63\\
  \bottomrule
\end{tabular}
\end{table}

\begin{table}[!htp]
  \centering
 \small 
  \caption{$R_{a}$ for different patch scales.}
  \label{rac}
  \begin{tabular}{cccc}
    \toprule
    \textbf{Scale} & \textbf{Monodepth2}& \textbf{Depthhints}& \textbf{Manydepth}\\
    \midrule
      0.1           &  0.97 &  0.95 & 0.97\\
      0.2           &  0.99  &  0.98 & 0.98 \\
      0.3           &  0.99 & 0.99  & 0.99\\
  \bottomrule
\end{tabular}
\end{table}

\section{Related Work}\label{sec:rw}

Unlike existing physical attacks that have targeted tasks like object detection \cite{Hu_ICCV, guesmi2023dap, advart}, image classification \cite{Athalye2018SynthesizingRA, ykholt_usenix, guesmi2023advrain}, and face recognition \cite{Sharif_CCS,Komkov_ICPR}, the area of attacks on depth estimation has received relatively limited attention. 

Yamanaka \cite{Yamanaka_Access} devises a method for generating printable adversarial patches, but the generated patch is trained to be applicable to random locations within the scene. On the other hand, Cheng \cite{Cheng_ECCV} concentrates on the inconspicuousness of the generated patch, ensuring that the patch remains unobtrusive and avoids drawing attention. However, the challenge with this patch lies in its object-specific nature, necessitating separate retraining for each target object. Furthermore, the patch's effectiveness is constrained by its limited affected region and its training for a specific context—namely, a fixed distance between the object and the camera—rendering it ineffective for varying distances.

Different from prior efforts, our emphasis lies in evaluating the comprehensive influence of the generated patch. We prioritize ensuring that the patch affects the entirety of the target object, thus guaranteeing a thorough deception of the DNN-based vision system. Our framework introduces a methodology that ensures the efficacy of the patch across various objects within the same class, accommodating distinct shapes and sizes. Importantly, our patch is designed to function across varying distances between the object's placement and the camera of the vision system.
\section{Conclusion}\label{sec:conclusion}

In this paper, we introduce a novel physical adversarial patch named SSAP, crafted with the explicit purpose of undermining MDE-based vision systems. SSAP distinguishes itself as an adaptive adversarial patch, demonstrating the capacity to fully hide objects or manipulate their perceived depth within a given scene, irrespective of their inherent size, shape, or placement. Our empirical investigations proves the effectiveness and resilience of our patch across diverse target objects. The achieved mean depth estimation error exceeding $50\%$, with over $99\%$ of the target region undergoing alteration. Furthermore, our patch exhibits durability against defense techniques grounded in input transformations.
The consequences of the proposed attack could result in significant harm, including loss, destruction, and endangerment to both life and property. Our findings serve as a clarion call to the research community, prompting the exploration of more robust and adaptive defense mechanisms.

\section*{Acknowledgment}

This research was partially funded by the NYUAD Center for Cyber Security (CCS), funded by Tamkeen under the NYUAD Research Institute Award G1104 and the Technology Innovation Institute (TII) under the "CASTLE: Cross-Layer Security for Machine Learning Systems IoT" project.

\tiny{
\bibliographystyle{IEEEtran}
\bibliography{bib}

\begin{thebibliography}{10}
\providecommand{\url}[1]{#1}
\csname url@samestyle\endcsname
\providecommand{\newblock}{\relax}
\providecommand{\bibinfo}[2]{#2}
\providecommand{\BIBentrySTDinterwordspacing}{\spaceskip=0pt\relax}
\providecommand{\BIBentryALTinterwordstretchfactor}{4}
\providecommand{\BIBentryALTinterwordspacing}{\spaceskip=\fontdimen2\font plus
\BIBentryALTinterwordstretchfactor\fontdimen3\font minus \fontdimen4\font\relax}
\providecommand{\BIBforeignlanguage}[2]{{%
\expandafter\ifx\csname l@#1\endcsname\relax
\typeout{** WARNING: IEEEtran.bst: No hyphenation pattern has been}%
\typeout{** loaded for the language `#1'. Using the pattern for}%
\typeout{** the default language instead.}%
\else
\language=\csname l@#1\endcsname
\fi
#2}}
\providecommand{\BIBdecl}{\relax}
\BIBdecl

\bibitem{Yamanaka_Access}
K.~Yamanaka, R.~Matsumoto, K.~Takahashi, and T.~Fujii, ``Adversarial patch attacks on monocular depth estimation networks,'' \emph{IEEE Access}, vol.~8, pp. 179\,094--179\,104, 2020.

\bibitem{Cheng_ECCV}
\BIBentryALTinterwordspacing
Z.~Cheng, J.~Liang, H.~Choi, G.~Tao, Z.~Cao, D.~Liu, and X.~Zhang, ``Physical attack on monocular depth estimation with optimal adversarial patches,'' 2022. [Online]. Available: \url{https://arxiv.org/abs/2207.04718}
\BIBentrySTDinterwordspacing

\bibitem{obstacles}
X.~Yang, J.~Chen, Y.~Dang, H.~Luo, Y.~Tang, C.~Liao, P.~Chen, and K.-T. Cheng, ``Fast depth prediction and obstacle avoidance on a monocular drone using probabilistic convolutional neural network,'' \emph{IEEE Transactions on Intelligent Transportation Systems}, vol.~22, no.~1, pp. 156--167, 2021.

\bibitem{autonomous2}
Y.~Wang, W.-L. Chao, D.~Garg, B.~Hariharan, M.~Campbell, and K.~Q. Weinberger, ``Pseudo-lidar from visual depth estimation: Bridging the gap in 3d object detection for autonomous driving,'' in \emph{2019 IEEE/CVF Conference on Computer Vision and Pattern Recognition (CVPR)}, 2019, pp. 8437--8445.

\bibitem{robotics1}
K.~Tateno, F.~Tombari, I.~Laina, and N.~Navab, ``Cnn-slam: Real-time dense monocular slam with learned depth prediction,'' \emph{2017 IEEE Conference on Computer Vision and Pattern Recognition (CVPR)}, pp. 6565--6574, 2017.

\bibitem{slam}
F.~Wimbauer, N.~Yang, L.~von Stumberg, N.~Zeller, and D.~Cremers, ``Monorec: Semi-supervised dense reconstruction in dynamic environments from a single moving camera,'' in \emph{2021 IEEE/CVF Conference on Computer Vision and Pattern Recognition (CVPR)}, 2021, pp. 6108--6118.

\bibitem{relocalization}
\BIBentryALTinterwordspacing
L.~von Stumberg, P.~Wenzel, N.~Yang, and D.~Cremers, ``Lm-reloc: Levenberg-marquardt based direct visual relocalization,'' \emph{CoRR}, vol. abs/2010.06323, 2020. [Online]. Available: \url{https://arxiv.org/abs/2010.06323}
\BIBentrySTDinterwordspacing

\bibitem{tesla2}
``Andrej karpathy - ai for full-self driving at tesla,'' \url{https://youtu.be/hx7BXih7zx8}, accessed: March 1, 2023.

\bibitem{tesla}
``Tesla ai day 2021,'' \url{https://www.youtube.com/live/j0z4FweCy4M?feature=share}, accessed: March 1, 2023.

\bibitem{toyota}
\BIBentryALTinterwordspacing
V.~Guizilini, R.~Ambrus, S.~Pillai, and A.~Gaidon, ``Packnet-sfm: 3d packing for self-supervised monocular depth estimation,'' \emph{CoRR}, vol. abs/1905.02693, 2019. [Online]. Available: \url{http://arxiv.org/abs/1905.02693}
\BIBentrySTDinterwordspacing

\bibitem{huawei}
\BIBentryALTinterwordspacing
S.~Aich, J.~M.~U. Vianney, M.~A. Islam, M.~Kaur, and B.~Liu, ``Bidirectional attention network for monocular depth estimation,'' \emph{CoRR}, vol. abs/2009.00743, 2020. [Online]. Available: \url{https://arxiv.org/abs/2009.00743}
\BIBentrySTDinterwordspacing

\bibitem{fu2018deep}
H.~Fu, M.~Gong, C.~Wang, K.~Batmanghelich, and D.~Tao, ``Deep ordinal regression network for monocular depth estimation,'' in \emph{Proceedings of the IEEE conference on computer vision and pattern recognition}, 2018, pp. 2002--2011.

\bibitem{song2021monocular}
M.~Song, S.~Lim, and W.~Kim, ``Monocular depth estimation using laplacian pyramid-based depth residuals,'' \emph{IEEE transactions on circuits and systems for video technology}, vol.~31, no.~11, pp. 4381--4393, 2021.

\bibitem{huynh2020guiding}
L.~Huynh, P.~Nguyen-Ha, J.~Matas, E.~Rahtu, and J.~Heikkil{\"a}, ``Guiding monocular depth estimation using depth-attention volume,'' in \emph{Computer Vision--ECCV 2020: 16th European Conference, Glasgow, UK, August 23--28, 2020, Proceedings, Part XXVI 16}.\hskip 1em plus 0.5em minus 0.4em\relax Springer, 2020, pp. 581--597.

\bibitem{chang2021transformer}
W.~Chang, Y.~Zhang, and Z.~Xiong, ``Transformer-based monocular depth estimation with attention supervision,'' in \emph{32nd British Machine Vision Conference (BMVC 2021)}, 2021.

\bibitem{varma2022transformers}
A.~Varma, H.~Chawla, B.~Zonooz, and E.~Arani, ``Transformers in self-supervised monocular depth estimation with unknown camera intrinsics,'' \emph{arXiv preprint arXiv:2202.03131}, 2022.

\bibitem{mimdepth}
H.~Chawla, K.~Jeeveswaran, E.~Arani, and B.~Zonooz, ``Image masking for robust self-supervised monocular depth estimation,'' 2023.

\bibitem{saam}
A.~Guesmi, M.~A. Hanif, B.~Ouni, and M.~Shafique, ``Saam: Stealthy adversarial attack on monocular depth estimation,'' \emph{IEEE Access}, vol.~12, pp. 13\,571--13\,585, 2024.

\bibitem{adam}
\BIBentryALTinterwordspacing
D.~P. Kingma and J.~Ba, ``Adam: A method for stochastic optimization,'' 2014. [Online]. Available: \url{https://arxiv.org/abs/1412.6980}
\BIBentrySTDinterwordspacing

\bibitem{Lin2014MicrosoftCC}
T.-Y. Lin, M.~Maire, S.~J. Belongie, J.~Hays, P.~Perona, D.~Ramanan, P.~Doll{\'a}r, and C.~L. Zitnick, ``Microsoft coco: Common objects in context,'' in \emph{ECCV}, 2014.

\bibitem{naturalistic}
Y.-C.-T. Hu, J.-C. Chen, B.-H. Kung, K.-L. Hua, and D.~S. Tan, ``Naturalistic physical adversarial patch for object detectors,'' in \emph{2021 IEEE/CVF International Conference on Computer Vision (ICCV)}, 2021, pp. 7828--7837.

\bibitem{mahendran2015understanding}
A.~Mahendran and A.~Vedaldi, ``Understanding deep image representations by inverting them,'' in \emph{Proceedings of the IEEE conference on computer vision and pattern recognition}, 2015, pp. 5188--5196.

\bibitem{monodepth2}
\BIBentryALTinterwordspacing
C.~Godard, O.~M. Aodha, and G.~J. Brostow, ``Digging into self-supervised monocular depth estimation,'' \emph{CoRR}, vol. abs/1806.01260, 2018. [Online]. Available: \url{http://arxiv.org/abs/1806.01260}
\BIBentrySTDinterwordspacing

\bibitem{Depthhints}
J.~Watson, M.~Firman, G.~Brostow, and D.~Turmukhambetov, ``Self-supervised monocular depth hints,'' in \emph{2019 IEEE/CVF International Conference on Computer Vision (ICCV)}, 2019, pp. 2162--2171.

\bibitem{Manydepth}
\BIBentryALTinterwordspacing
J.~Watson, O.~M. Aodha, V.~Prisacariu, G.~Brostow, and M.~Firman, ``The temporal opportunist: Self-supervised multi-frame monocular depth,'' in \emph{2021 IEEE/CVF Conference on Computer Vision and Pattern Recognition (CVPR)}.\hskip 1em plus 0.5em minus 0.4em\relax Los Alamitos, CA, USA: IEEE Computer Society, jun 2021, pp. 1164--1174. [Online]. Available: \url{https://doi.ieeecomputersociety.org/10.1109/CVPR46437.2021.00122}
\BIBentrySTDinterwordspacing

\bibitem{kitti}
A.~Geiger, P.~Lenz, and R.~Urtasun, ``Are we ready for autonomous driving? the kitti vision benchmark suite,'' in \emph{2012 IEEE Conference on Computer Vision and Pattern Recognition}, 2012, pp. 3354--3361.

\bibitem{casia}
S.~Yu, D.~Tan, and T.~Tan, ``A framework for evaluating the effect of view angle, clothing and carrying condition on gait recognition,'' in \emph{18th international conference on pattern recognition (ICPR'06)}, vol.~4.\hskip 1em plus 0.5em minus 0.4em\relax IEEE, 2006, pp. 441--444.

\bibitem{Hu_ICCV}
Y.-C.-T. Hu, J.-C. Chen, B.-H. Kung, K.-L. Hua, and D.~S. Tan, ``Naturalistic physical adversarial patch for object detectors,'' in \emph{2021 IEEE/CVF International Conference on Computer Vision (ICCV)}, 2021, pp. 7828--7837.

\bibitem{guesmi2023dap}
A.~Guesmi, R.~Ding, M.~A. Hanif, I.~Alouani, and M.~Shafique, ``Dap: A dynamic adversarial patch for evading person detectors,'' in \emph{Proceedings of the IEEE/CVF Conference on Computer Vision and Pattern Recognition (CVPR)}, June 2024, pp. 24\,595--24\,604.

\bibitem{advart}
\BIBentryALTinterwordspacing
A.~Guesmi, I.~M. Bilasco, M.~Shafique, and I.~Alouani, ``Advart: Adversarial art for camouflaged object detection attacks,'' 2024. [Online]. Available: \url{https://arxiv.org/abs/2303.01734}
\BIBentrySTDinterwordspacing

\bibitem{Athalye2018SynthesizingRA}
A.~Athalye, L.~Engstrom, A.~Ilyas, and K.~Kwok, ``Synthesizing robust adversarial examples,'' in \emph{ICML}, 2018.

\bibitem{ykholt_usenix}
K.~Eykholt, I.~Evtimov, E.~Fernandes, B.~Li, A.~Rahmati, F.~Tram\`{e}r, A.~Prakash, T.~Kohno, and D.~Song, ``Physical adversarial examples for object detectors,'' in \emph{Proceedings of the 12th USENIX Conference on Offensive Technologies}, ser. WOOT'18.\hskip 1em plus 0.5em minus 0.4em\relax USA: USENIX Association, 2018, p.~1.

\bibitem{guesmi2023advrain}
A.~Guesmi, M.~A. Hanif, and M.~Shafique, ``Advrain: Adversarial raindrops to attack camera-based smart vision systems,'' \emph{Information}, vol.~14, no.~12, p. 634, 2023.

\bibitem{Sharif_CCS}
\BIBentryALTinterwordspacing
M.~Sharif, S.~Bhagavatula, L.~Bauer, and M.~K. Reiter, ``Accessorize to a crime: Real and stealthy attacks on state-of-the-art face recognition,'' in \emph{Proceedings of the 2016 ACM SIGSAC Conference on Computer and Communications Security}, ser. CCS '16.\hskip 1em plus 0.5em minus 0.4em\relax New York, NY, USA: Association for Computing Machinery, 2016, p. 1528–1540. [Online]. Available: \url{https://doi.org/10.1145/2976749.2978392}
\BIBentrySTDinterwordspacing

\bibitem{Komkov_ICPR}
\BIBentryALTinterwordspacing
S.~Komkov and A.~Petiushko, ``Advhat: Real-world adversarial attack on arcface face id system,'' in \emph{2020 25th International Conference on Pattern Recognition (ICPR)}.\hskip 1em plus 0.5em minus 0.4em\relax Los Alamitos, CA, USA: IEEE Computer Society, jan 2021, pp. 819--826. [Online]. Available: \url{https://doi.ieeecomputersociety.org/10.1109/ICPR48806.2021.9412236}
\BIBentrySTDinterwordspacing

\end{thebibliography}
}
\end{document}